\documentclass{article}

\usepackage[left=1in,right=1in,top=1in,bottom=1in,%
            footskip=.25in]{geometry}

\usepackage[utf8]{inputenc} 
\usepackage[T1]{fontenc}
\usepackage{url}
\usepackage{ifthen}
\usepackage{cite}

\usepackage{float}
\usepackage{subfloat}

\usepackage{mathtools,amssymb,graphicx,times,comment}
\usepackage{caption}
\usepackage{subcaption}
\usepackage{amsmath}
\usepackage{amsthm}
\usepackage{bbm}
  {

  }
 
\usepackage{tcolorbox}
\usepackage[hidelinks]{hyperref}
\definecolor{darkred}{RGB}{150,0,0}
\definecolor{darkgreen}{RGB}{0,150,0}
\definecolor{darkblue}{RGB}{0,0,150}
\hypersetup{colorlinks=true, linkcolor=red, citecolor=darkgreen, urlcolor=darkblue}

\usepackage[font=small,labelfont=bf]{caption}

  \usepackage{breqn}
  
\newcommand{\mup}{\nu}

\newcommand{\betabs}{\mathbf{\betab}_\star}
\newcommand{\yh}{\hat{y}}
\newcommand{\Rce}{\widehat{\mathcal{R}}_{\rm emp}}

\newcommand{\Rct}{\mathcal{R}_{\rm train}}

\newcommand{\Ecsep}{\Ec_{\rm{sep}}}

\newcommand{\Rad}[1]{{\rm{Rad}}\left(#1\right)}

\newcommand{\mathleft}{\@fleqntrue\@mathmargin0pt}
\newcommand{\mathcenter}{\@fleqnfalse}

\newcommand{\rhod}{f}

\newcommand{\R}{\mathbb{R}}

    \newcommand{\gammab}{\boldsymbol\gamma}

\newcommand{\betabh}{\boldsymbol{\widehat{\beta}}}

\newcommand{\rP}{\stackrel{P}{\rightarrow}}

\DeclarePairedDelimiterX{\inp}[2]{\langle}{\rangle}{#1, #2}

\newcommand{\Id}{\mathbf{I}}

\newcommand{\etab}{\boldsymbol{\eta}}

\usepackage{dsfont}

\newcommand{\simiid}{\stackrel{\text{iid}}{\sim}}

\newcommand{\Pro}{\mathbb{P}}

\theoremstyle{theorem}
\newtheorem{propo}{Proposition}[section]
\theoremstyle{remark}
\newtheorem{remark}{Remark}
\theoremstyle{definition}

\newcommand{\eps}{\varepsilon}

\newcommand{\sign}{\mathrm{sign}}

\newcommand{\E}{\mathbb{E}}                    

\newcommand{\nn}{\notag}
\newcommand{\Rb}{\mathbb{R}}

\newcommand{\W}{\mathbf{W}}

\newcommand{\x}{\mathbf{x}}

\newcommand{\w}{\mathbf{w}}

\newcommand{\y}{\mathbf{y}}

\newcommand{\z}{\mathbf{z}}

\newcommand{\betab}{\boldsymbol{\beta}}

\newcommand{\Tc}{{\mathcal{T}}}
\newcommand{\Sc}{{\mathcal{S}}}

\newcommand{\Rc}{\mathcal{R}}
\newcommand{\Nn}{\mathcal{N}}

\newcommand{\Ec}{\mathcal{E}}

\newcommand{\beq}{\begin{equation}}
\newcommand{\eeq}{\end{equation}}
\newcommand{\bea}{\begin{align}}
\newcommand{\eea}{\end{align}}

\newcommand{\vp}{\vspace{4pt}}

\def\bea#1\eea{\begin{align}#1\end{align}}

\title{Analytic Study of Double Descent \\in Binary Classification: The Impact of Loss}

\author{
Ganesh Kini  and  Christos Thrampoulidis \\
University of California, Santa Barbara, Department of Electrical and Computer Engineering.
}

\begin{document}

\maketitle

\begin{abstract}
Extensive empirical evidence reveals that, for a wide range of different learning methods and datasets, the risk curve exhibits a double-descent (DD) trend as a function of the model size. In our recent coauthored paper \cite{Zeyu}, we studied binary linear classification models and showed that the test error of gradient descent (GD) with \emph{logistic} loss undergoes a DD. In this paper, we complement these results by extending them to GD with \emph{square} loss. We show that the DD phenomenon persists, but we also identify several differences compared to logistic loss. This emphasizes that crucial features of DD curves (such as their transition threshold and global minima) depend both on the training data and on the learning algorithm. We further study the dependence of DD curves on the size of the training set. Similar to our earlier work, our results are analytic: we plot the DD curves by first deriving sharp asymptotics for the test error under Gaussian features. Albeit simple, the models permit a principled study of DD features, the outcomes of which theoretically corroborate related empirical findings occurring in more complex learning tasks. 

%

\end{abstract}


%
\section{Introduction}

\subsection{Motivation}
It is common practice in modern learning architectures to choose the number of training parameters such that it overly exceeds the size of the training set. Despite the \emph{overparametrization}, such architectures are known to generalize well in practice. 
Specifically, it has been consistently reported, in the recent literature, that the risk curve as a function of the model size exhibits a, so called, ``\emph{double-descent}" (DD) behavior \cite{belkin2018reconciling}. An initial descent is followed by an ascend as predicted by conventional statical wisdom of the approximation-generalization tradeoff, but as the model size surpasses a certain transition threshold a second descent occurs. This phenomenon has been 
%
demonstrated experimentally for decision trees, random features and two-layer neural networks (NN) in \cite{belkin2018reconciling} and, more recently, for deep NNs (including ResNets, standard CNNs and Transformers) in \cite{DDD}; see also \cite{spigler2019jamming,geiger2019scaling} for similar observations.


Our work is motivated by these empirical findings and has a two-fold goal. First, we demonstrate analytically that the DD phenomenon is already present in certain simple binary linear classification settings. Second, for these simple settings, we undertake a principled study of how certain features of DD curves vary depending  on the data, as well as on the training algorithms. The outcomes of this study theoretically corroborate related empirical findings occurring in more complicated learning methods for a variety of learning tasks. Most other recent efforts towards theoretically understanding DD curves focus on linear regression. Instead, we study binary linear classification; see Section \ref{sec:rel} for a comparison to related works.

\begin{figure}
	\centering
		\centering
		\includegraphics[width=.6\linewidth]{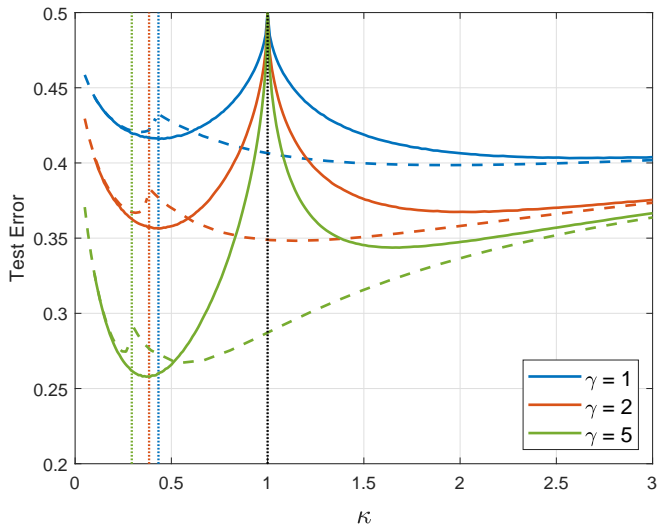}
\caption{The effect of the training algorithm (square- vs logistic- loss) on the DD curves of the test error as a function of the overparametrization ratio $\kappa$ for logistic data under a polynomial feature selection model (see text for details). In this paper, we derive the risk curves for square loss, which are plotted in solid lines. The curves for logistic-loss, shown in dashed lines, are derived in \cite{Zeyu}. The vertical lines indicate the phase-transition threshold for each case. For square-loss, the threshold is always at $1$ (after which data can be linearly interpolated), but for logistic loss its value $\kappa_{\star,\rm LOG}$ (after which data can be linearly separable) depends on the data. Note that the test error of logistic loss is less sensitive to variations of the model size. However, there are regimes where the square loss achieves better classification performance. The different values of parameter $\gamma$ play the role of SNR for fixed model size; see text for details.}
\label{fig:log_data_sq_loss_log_loss}
\end{figure}

\subsection{Contributions}
We outline the paper's contributions. 
\begin{enumerate}
\item First, we evaluate the classification error of gradient descent (GD) with squares loss for two simple, yet popular, models for binary linear classification, namely logistic and Gaussian mixtures (GM) models. Specifically, we study a linear high-dimensional regime in which the size $n$ of training set and the model size $p$ both grow large at a proportional rate $\kappa:=p/n$. We derive asymptotic formulae for all values of $\kappa>0$. 
\item Second, we use the theoretical predictions to show that the test error of certain simple binary classification models undergoes a double descent (DD) behavior when performing GD on square-loss. 
\item The DD curves that we obtain are qualitatively different than corresponding curves for logistic loss derived in \cite{Zeyu}. Specifically, we analytically show that the characteristics of the DD curve, such as the location and the shape of its kink, depend not only on the data, but also on the training algorithm (aka, logistic vs square loss). Our results identify SNR and data model regimes in which the the test error achieved by LS optimized over the model size $p$ is \emph{lower} than the corresponding error for logistic loss. However, the error of logistic loss is less sensitive to variations of the model size. See Figure \ref{fig:log_data_sq_loss_log_loss}.

\item Finally, motivated by corresponding empirical findings recently reported in \cite{DDD}, we use our results to analytically study the dependence of the DD curves on the size of the training set. In resemblance to \cite{DDD}, we show that, because of the inherent W-shape of the curve, there exist ``bad" choices of the model size $p$ such that the test error of GD increases despite the larger sample-size $n$. See Figure \ref{fig:varying_n}. 
\end{enumerate}

Overall, our study theoretically demonstrates that several key aspects of recent empirical findings on DD behavior of overparametrized learning architectures are already present in simple binary classification models and linear classification algorithms. 
\subsection{Related works}\label{sec:rel}

Recent efforts towards theoretically understanding the phenomenon of double descent (DD) focus on linear regression settings \cite{advani2017high,hastie2019surprises,muthukumar2019harmless,belkin2019two,xu2019many,bartlett2019benign,nakkiran2019more,mei2019generalization}; see \cite[App.~C]{DDD} for an extended discussion on their individual contributions. Out of these, the closest related to our paper are \cite{hastie2019surprises,belkin2019two}, which use random matrix theory (RMT) to study DD curves of gradient descent (GD) for linear regression with square loss in certain simple mis-match models with Gaussian features. Here, we extend these results to classification. To the best of our knowledge, the only previous works that study DD in a classification setting are \cite{montanari2019generalization} and our co-authored paper \cite{Zeyu}, specifically, both papers compute asymptotics of GD with logistic loss. In this paper, we extend the results of \cite{Zeyu} to GD with square loss, thus emphasizing the dependence of DD not only on the data, but also on the training algorithm. 



On a technical level,  our asymptotic analysis fits in the rapidly growing recent literature on sharp performance guarantees of convex optimization based estimators; see \cite{montanariLASSO,StoLasso,COLT,montanari13,Master,karoui15,wang2019does} and many references therein. Specifically, we utilize the convex Gaussian min-max theorem (CGMT) \cite{COLT}. Most of the aforementioned works study linear regression, for which the CGMT framework has been shown to be powerful and applicable under several variations; e.g., \cite{Master,celentano2019fundamental,abbasi2019universality} and references therein.
In contrast, our results hold for binary classification. For the derivations we utilize the machinery recently put forth in \cite{svm_abla,taheri2019sharp,Zeyu,abbasi2019universality}, which demonstrate that the CGMT can be also applied to classification problems. Closely related ideas were previously introduced in \cite{NIPS,PhaseLamp}. Here, we introduce necessary adjustments to accommodate the needs of the specific data generation model and focus on classification error. There are several other works on sharp asymptotics of binary linear classification both for the logistic \cite{candes2018phase,sur2019modern,salehi2019impact} and the Gaussian-mixture model \cite{mai2019large,logistic_regression,huang2017asymptotic}. While closely related, these works differ in terms of motivation, model specification, end-results, and proof techniques. 

\section{Learning model}\label{sec:model}


\subsection{Two data models}\label{sec:data}

Let $\x\in\R^d$ denote the feature vector and $y\in\{\pm 1\}$ denote the class label. We study supervised binary classification under the following two popular data models (e.g., \cite[Sec.~3.1]{williams2006gaussian}).

\vp
\noindent\emph{\textbf{A generative Logistic model}:}~We model the marginal $p(y=+1|\x)$ with the sigmoid function $f(\x^T\etab_0)=(1+e^{-\x^T\etab_0})^{-1}$, where $\etab_0\in\Rb^d$ is an unknown weight vector. Also, we assume IID Gaussian feature vectors $\x$. 
%
For compactness let $\Rad{p}$ denote a symmetric Bernoulli distribution with probability $p$ for the value $+1$ and probability $1-p$ for the value $-1$. Then, the logistic model with Gaussian features is:
\bea\label{eq:yi_log}
y\sim\Rad{\rhod(\x^T\etab_0)},~~~~\x\sim\Nn(0,\Id_d).
\eea


\vp
\noindent\emph{\textbf{A discriminative Gaussian mixtures (GM) model}:}~We model the class-conditional densities $p(\x|y)$ with Gaussians.  Specifically, each data point belongs to one of two classes \{$\pm$1\} with probabilities $\pi_{\pm1}$ such that $\pi_{+1} + \pi_{-1}=1$. If it comes from class \{$\pm$1\} (so that $y=\pm 1$) , then the feature vector $\x\in\R^d$ is an IID Gaussian vector with mean $\pm\etab_0\in\R^d$. In short:
\bea\label{eq:yi_GM}
y = \pm 1 \Leftrightarrow \x\sim\Nn\left(\pm\etab_0,\Id_d\right).
\eea


\subsection{Training-data: A mismatch model}\label{sec:train_model}
 During training, we assume access to $n$ data points generated according to either \eqref{eq:yi_log} or \eqref{eq:yi_GM}. 
We allow for the possibility that only a subset $\Sc\subset[d]$ of the total number of features is known during training.
Concretely, for each feature vector $\x_i,~i\in[n]$, the learner only knowns a sub-vector $\w_i:=\x_i(\Sc):=\{\x_i(j),~j\in\Sc\},$
for a chosen set $\Sc\subset[d]$.
We denote the size of the \emph{known} feature sub-vector as $p:=|\Sc|$. Onwards, we choose $\Sc=\{1,2,\ldots,p\}$ \footnote{This assumption is without loss of generality for our asymptotic model specified in Section \ref{sec:setting}.}, i.e., select the features sequentially in the order in which they appear.

 Clearly, $1\leq p\leq d$. Overall, the training set $\Tc(\Sc)$ consists of $n$ data pairs:
\begin{align}\label{eq:TS}
&\quad\Tc(\Sc):=\left\{(\w_i,y_i), i=1,\dots,n\right\},\qquad \\&\w_i=\x_i(\Sc)\in\R^p~\text{ where }~(y_i,\x_i)\sim\eqref{eq:yi_log}~\text{or}~\eqref{eq:yi_GM}\nn.
\end{align}
When clear from context, we simply write 
$
\Tc\sim\eqref{eq:yi_log},
$
or $\Tc\sim\eqref{eq:yi_GM}$ 
if the training set $\Tc$ is generated according to \eqref{eq:yi_log} or \eqref{eq:yi_GM}, respectively. We let $\W\in\R^{n\times p}$ denote the \emph{feature matrix}, i.e., $\W = [\w_1,\w_2,\ldots,\w_n]^T$.




\subsection{Classification algorithm}\label{sec:class}
 Having access to the training set $\Tc(\Sc)$, the learner obtains an estimate $\betabs\in\R^p$ of the weight vector $\etab_0\in\R^d$. For a newly generated sample $(\x,y)$ (and $\w:=\x(\Sc)$), 
she forms a \emph{linear classifier} and decides a label $\yh$ for the new sample as:
$\yh = \sign(\w^T\betabs).$
%
The estimate $\betabs$ is  minimizing the empirical risk 
\begin{align}\label{eq:Remp}
\min_\betab~\Rce(\betab):=\frac{1}{n}\sum_{i=1}^n\ell(y_i\w_i^T\betab),
\end{align}
 for certain loss function $\ell:\R\rightarrow\R$. A common approach to minimize $\Rce$ is by constructing gradient descent (GD) iterates $\betab_{(k)},~k\geq0$ via $$
\betab_{(k+1)} = \betab_{(k)} - \eta_k\nabla\Rce(\betab_{(k)}),
$$
for step-size $\eta_k>0$ and arbitrary $\betab_{(0)}$. We run GD until convergence and set
$\betabs=\lim_{k\rightarrow\infty}\betab_{(k)}.$
We focus on the following two popular choices for the loss function $\ell$:

\begin{itemize}
\item \textbf{Logistic:}  $\ell(t):=\log\big(1+e^{-t}\big),$
\item \textbf{Least-squares:}  $\ell(t):=(1-t)^2.$
\end{itemize}

%

\noindent For the least-squares loss, note that $(y_i\w_i^T\betab-1)^2=(y_i-\w_i^T\betab)^2$ since $y_i\in\{\pm 1\}$.

%
%

\vp
For a new sample $(\x,y)$ we measure test error of $\hat{y}$ by the expected \emph{risk} with respect to the 0-1 loss
\begin{align}\label{eq:risk}
\Rc(\betabs) := \E\left[\mathds{1}\big(\yh\neq y \big)\right].
\end{align}
The expectation is over the distribution of the \emph{new} data sample $(\x,y)$ generated according to either \eqref{eq:yi_log}~\text{or}~\eqref{eq:yi_GM}. In particular, note that $\Rc(\betabs)$ is a function of the training set $\Tc$. 
\noindent Finally, the training error of $\betab_\star$ is given by
\begin{align}\label{eq:train_error}
\Rct(\betabs):=\frac{1}{n}\sum_{i=1}^n \mathds{1}\big(\yh\neq y_i \big).
\end{align}


\subsection{Optimization regimes of Gradient Descent}\label{sec:GD}

Our analysis of the classification performance of Gradient Descent (GD) makes use of the following characterizations of the converging behavior of GD iterations for a given training set $\Tc$ and loss function $\ell$ (logistic and LS).


\vp
\noindent{\textbf{Logistic loss:}}~For the logistic loss the convergence behavior depends critically on the event that the training data are linearly separable, i.e., 
$
\Ecsep := \left\{ ~\exists \betab\in\R^p~:~y_i(\w_i^T\betab)\geq1,~\forall i\in[n]~ \right\},
$
\cite{pmlr-v99-ji19a,soudry2018implicit,telgarsky2013margins}.
 On the one hand, when $\Ecsep$ does \emph{not} hold, then $\Rce(\beta)$ is coercive, its sub-level sets are closed and GD iterates converge to a finite minimizer $\betabh$ of the empirical loss in \eqref{eq:Remp}. Furthermore, when the feature matrix is full column rank, then the loss is strongly convex and $\betabh$ is unique. On the other hand,  when data are separable, the set of minimizers of the empirical loss is unbounded. In this case, \cite{pmlr-v99-ji19a} shows that the normalized iterations of GD converge to the \emph{max-margin classifier}, i.e.,  $\big\|\frac{\betab_{(k)}}{\|\betab_{(k)}\|_2}-\frac{\betabh}{\|\betabh\|_2}\big\|_2\stackrel{k\rightarrow\infty}{\longrightarrow} 0,$ where 
\bea\label{eq:SVM}
\betabh=\arg\min_{\betab}~\|\betab\|_2~~\text{sub. to}~~~ y_i\w_i^T\betab\geq 1, \forall i.
\eea

\vp
\noindent{\textbf{Least-Squares:}}~For least-squares loss, it is a well-known fact  that for appropriately small and constant step size the GD iterations converge to the {min-norm least-squares regression solution} (e.g., see \cite[Prop.~1]{hastie2019surprises}):
$\betabh=\arg\min\big\{ \|\betab\|_2~|~ \betab ~\text{minimizes}~ \|\y-\W\betab\|_2 \big\}.$
Specifically, when the feature matrix $\W$ is full column-rank, then the empirical loss is strongly convex and GD converges to $\betabh=(\W^T\W)^{-1}\W^T\y$. In contrast, when $\W$ is full row-rank, then out of all the solutions $\betab$ that linearly interpolate the data, i.e., $\y=\W\betab$, GD converges to the \emph{min-norm regression solution}:
\bea\label{eq:min-norm}
\betabh=\arg\min_{\betab}~\|\betab\|_2~~\text{sub. to}~~~ y_i = \w_i^T\betab, \forall i.
\eea

\section{Sharp Asymptotics}\label{sec:theory}

We present sharp asymptotic formulae for the classification error of GD iterations for the logistic and least-squares (LS) losses under the learning model of Sections \ref{sec:data} and \ref{sec:train_model}, as well as, the linear asymptotic setting presented in Section \ref{sec:setting}. The first key step of our analysis, makes use of the convergence results discussed in Section \ref{sec:GD} that relate GD for logistic and LS loss to the convex programs \eqref{eq:Remp}, \eqref{eq:SVM} and \eqref{eq:min-norm} depending on whether certain deterministic properties of the training data $\Tc$ (such as linear separability or conditioning of the feature matrix) hold. In Section \ref{sec:PT}, we show that in the linear asymptotic regime with Gaussian features the change in behavior of convergence of GD undergoes sharp phase-transitions for both loss functions and data models (logistic and Gaussian-mixtures). The boundary of the phase-transition separates the so-called under- and over-parametrized regimes. Thus, for both logistic and LS loss and each one of its two corresponding regimes, GD converges to an associated convex program. The second key-step of our analysis, uses the Convex Gaussian min-max Theorem (CGMT) \cite{COLT,Master} to evaluate the classification error of these convex programs.

This paper focuses on LS loss. Corresponding results for logistic loss are derived in the coauthored work \cite{Zeyu}. In Section \ref{sec:num}, we combine the results from both papers to compare DD curves of logistic and LS loss. In this section, we only present the asymptotic formulae for the case of LS; please see \cite[Sec. 3.3 \& 3.4]{Zeyu} for the case of logistic loss.


\subsection{Asymptotic setting}\label{sec:setting}

Our asymptotic results hold in a \emph{linear asymptotic regime} where $n,d,p\rightarrow+\infty$ such that 
\bea\label{eq:linear}
{d/n}\rightarrow \zeta>0 \quad \text{and}\quad  {p/n}\rightarrow \kappa\in(0,\zeta],
\eea
where recall the following definitions:

\indent$\bullet$~ $d$: dimension of the ambient space,\\
\indent$\bullet$~ $n$: training sample size,\\
\indent$\bullet$~ $p$:  model size.

\noindent To quantify the effect of the overparametrization ratio $\kappa$ on the test error, we decompose the feature vector $\x_i\in\R^d$ to its \emph{known} part $\w_i\in\R^p$ and to its unknown part $\z_i$:
$
\x_i:=[\w_i^T,\z_i^T]^T.\nn
$
Then, we let $\betab_0\in\R^p$ (resp., $\gammab_0\in\R^{d-p}$) denote the vector of weight coefficients corresponding to the known (resp., unknown) features such that 
$
\etab_0:=[\betab_0^T,\gammab_0^T]^T.
$
 In this notation, we study a sequence of problems of increasing dimensions $(n,d,p)$ as in \eqref{eq:linear} that further satisfy:
\bea
\|\etab_0\|_2 \rightarrow r~\text{ and }~ \label{eq:strengths} \|\betab_0\|_2 \rightarrow s:=s(\kappa).
\eea
%
%
The parameter $s\in(0,r]$ can be thought of as the useful signal strength. Our notation specifies that $s(\kappa)$ is a function of $\kappa$. We are interested in functions $s(\kappa)$ that are increasing in $\kappa$ such that the signal strength increases as more features enter into the training model; see Sec.~\ref{sec:explicit} for explicit parameterizations. 
For each triplet $(n,d,p)$ in the sequence of problems that we consider, the corresponding training set $\Tc=\Tc(\{1,2,\ldots,p\})$ follows \eqref{eq:TS}.

\subsection{Phase-transitions of optimization regimes}\label{sec:PT}
As discussed in Section \ref{sec:model}, the behavior of GD with square loss changes depending on whether the feature matrix $\W$ is full column- or row-rank. On the one hand, when $p>n$ (aka $\kappa>1$) and $\W\W^T$ is invertible, then GD converges to the unique LS solution. On the other hand, when $n>p$ (aka $\kappa<1$) and $\W^T\W$ is invertible, then the data can be linearly interpolated and GD converges to the min-norm solution in \eqref{eq:min-norm}. Under the Gaussian feature model, the following well-known result shows that the aforementioned change in behavior undergoes a sharp phase-transition with transition threshold $1$. 
\begin{propo}[Linear interpolation threshold (e.g. \cite{VerBook})]\label{propo:PT} 
For a training set $\Tc$ that is generated by either of the two models in \eqref{eq:yi_log} or \eqref{eq:yi_GM}, the following hold:
\bea
\nn \kappa>1~&\Rightarrow~\lim_{n,p,d\rightarrow\infty}\Pr\left(~ \W\W^T \text{ is invertible } ~\right)=1,\\
\nn \kappa<1~&\Rightarrow~\lim_{n,p,d\rightarrow\infty}\Pr\left(~ \W^T\W \text{ is invertible } ~\right)=1.
\eea
\end{propo}
\noindent Put in words: when $\kappa>1$, with probability 1, the LS empirical loss can be driven to zero and GD converges to the min-norm solution $\betabh$ of \eqref{eq:min-norm}. Moreover, since $\forall i: \y_i\x_i^T\betabh=1$, the training error $\Rct(\betabh)$ is also going to zero in the regime $\kappa>1$.

\begin{remark}[Comparison to Logistic loss]  In \cite[Prop.~3.1]{Zeyu}, the authors prove a corresponding phase-transition for the case of logistic loss (see also \cite{candes2018phase}, \cite{montanari2019generalization}). Specifically, for both the Logistic and GM models we compute corresponding thresholds $\kappa_{\star,\rm{LOG}}\in(0,1/2)$ and $\kappa_{\star,\rm{GM}}\in(0,1/2)$, such that GD with logistic loss converges to:  (i) the unique solution of \eqref{eq:Remp} for $\kappa$ less than the threshold; (ii) to the max-margin solution \eqref{eq:SVM} for $\kappa$ larger than the threshold. In the latter case, the training data are separable; hence, both the empirical logistic loss and the training error are zero with probability one. According to Proposition \ref{propo:PT}, the corresponding threshold for square loss is $1$ for both the logistic and GM models. Note that this is strictly larger (in fact, at least double) than both $\kappa_{\star,\rm{LOG}}$ and $\kappa_{\star,\rm{GM}}$. Moreover, both $\kappa_{\star,\rm{LOG}}$ and $\kappa_{\star,\rm{GM}}$ are functions of the SNR $r$, which is \emph{not} the case for the square loss.
\end{remark}

%
%

%
%

\subsection{High-dimensional asymptotics}\label{sec:asy}

For each increasing triplet $(n,d,p)$ and sequence of problem instances following the asymptotic setting of Section \ref{sec:setting}, let $\betabh\in\R^d$ be the sequence of vectors of increasing dimension corresponding to the converging point of GD for square loss. The Propositions \ref{propo:log} and \ref{propo:GM} below characterize the asymptotic value of the classification error of the sequence of $\betabh$'s for the logistic and Gaussian-mixture (GM) models, respectively.

 
 In what follows, for a sequence of random variables $\mathcal{X}_{n,p,d}$ that converges in probability to a constant $c$ in the limit of \eqref{eq:linear}, we simply write $\mathcal{X}_{n,p,d}\rP c$.  Furthermore, let $Q(t)=\frac{1}{\sqrt{2\pi}}\int_{t}^{+\infty}e^{-t^2/2}\mathrm{d}t$ be the Gaussian Q-function.

\begin{propo}[Logistic model]\label{propo:log}
Fix total signal strength $r$, overparametrization ratio $\kappa>0$ and parameter $s=s(\kappa)\in(0,r]$. With these, consider a training set $\Tc$ that is generated by the Logistic model in \eqref{eq:yi_log} and satisfies \eqref{eq:strengths}. Define the following random variables:  $$H,G,Z\simiid\Nn(0,1), ~~ Y\sim\Rad{\rhod(s\,G+\sqrt{r^2-s^2}\,Z)},$$ and denote, 
\bea
\mup := 2\E_H\left[\frac{\exp(-rH)}{(1+\exp(-rH))^2}\right].\nonumber
\eea
Finally, define the effective noise parameter $\rho:={\alpha}/{\mu}$, where $\mu$ and $\alpha>0$ are computed as follows:
\bea\label{eq:mu_log}
\mu = \begin{cases}
s\mup,&\text{if}~\kappa<1,\\
{s\mup}/{\kappa},&\text{if}~\kappa>1.
\end{cases},
\quad\quad
%
\alpha^2 = \begin{cases}
(1-s^2\mup^2)\frac{\kappa}{1-\kappa},&\text{if}~\kappa<1,\\
\frac{\kappa^2 + (1-2\kappa)s^2\mup^2}{\kappa^2(\kappa-1)},&\text{if}~\kappa>1.
\end{cases}
\eea
Then, $$\Rc(\betabh)\rP\Pro\left(\rho\,H+GY<0\right).\nn$$
\end{propo}

\begin{propo}[GM model]\label{propo:GM}
Fix total signal strength $r$, overparametrization ratio $\kappa>0$ and parameter $s=s(\kappa)\in(0,r]$. With these, consider a training set $\Tc$ that is generated by the GM model in \eqref{eq:yi_GM} and satisfies \eqref{eq:strengths}. Define the effective noise parameter $\rho:={\alpha}/{\mu}$, where $\mu$ and $\alpha>0$ are computed as follows:
\bea\label{eq:mu_log}
\mu = \begin{cases}
\frac{s}{1+s^2},&\text{if}~\kappa<1,\\
\frac{s}{\kappa+s^2},&\text{if}~\kappa>1.
\end{cases},
%
\quad\quad
\alpha^2:= \begin{cases}
\frac{\kappa}{1-\kappa}\frac{1}{1+s^2},&\text{if}~\kappa<1,\\
\frac{\kappa^2 + s^2}{(\kappa-1)(\kappa+s^2)^2},&\text{if}~\kappa>1.
\end{cases}
\eea
Then, $$\Rc(\betabh)\rP Q\Big(\frac{s}{\sqrt{1+\rho^2}}\Big).$$ 
\end{propo}

In order to prove Propositions \ref{propo:log} and \ref{propo:GM}, we use the fact that (cf. Sec. \ref{sec:PT}): (i) for $\kappa<1$, $\betabh=\arg\min_{\betab}\|\y-\W\betab\|_2$, where the minimizer is unique; (ii) for $\kappa>1$, $\betabh$ is as in \eqref{eq:min-norm}. Thus, it suffices to evaluate the asymptotics of these two convex programs, which we do by using the CGMT \footnote{Note that both convex minimization programs have closed form solutions in terms of the Moore-Penrose pseudoinverse $(\W^T\W)^\dagger$, i.e., $\betabh=(\W^T\W)^\dagger\W^T\y$. Thus, it is also possible to use random matrix theory tools to analyze the classification performance of $\betabh$. We find the CGMT more direct to implement; it also provides a unified treatment with the results in \cite{Zeyu}.}. The machinery follows closely (in fact, is simpler) the proofs for logistic loss that were presented in \cite{Zeyu}. 
To the best of our knowledge, the formulae presented in Propositions \ref{propo:log} and \ref{propo:GM} are novel, with the exception of the result of Proposition  \ref{propo:log} for $\kappa<1$, which follows from \cite{NIPS,taheri2019sharp}.


\begin{remark}[Critical regime around the peak] For both the logistic and GM models it can be checked that as $\kappa$ approaches the linear interpolation threshold $1$ from either left or right: $\rho\rightarrow+\infty$. Thus, irrespective of $r$ and $s$, in the limit of $\kappa=1$ (i.e., when the model size $p$ equals the training size $n$), the test error $\Rc$ converges to $1/2$ and there is no predictive ability for classification. 
\end{remark}

\begin{remark}[Other metrics] Our analysis further predicts the asymptotic behavior of other performance metrics of interest. For example, the $\ell_2$-norm of the trained model $\|\betabh\|_2$ converges to $\sqrt{\mu^2+\alpha^2}$ for the same values as in Propositions \ref{propo:log} and \ref{propo:GM}. Moreover, it can be shown \cite{Zeyu,AISTATS2020} that $\|\betabh-\frac{\mu}{s}\betab_0\|_2^2 \rP \alpha^2$. Thus, the estimated $\betabh$ is centered around $\frac{\mu}{s}\betab_0$ and $\alpha^2$ captures the deviations around it.
\end{remark}
\begin{figure}
	\centering
	\begin{subfigure}{.48\textwidth}
		\centering
		\includegraphics[width=\linewidth]{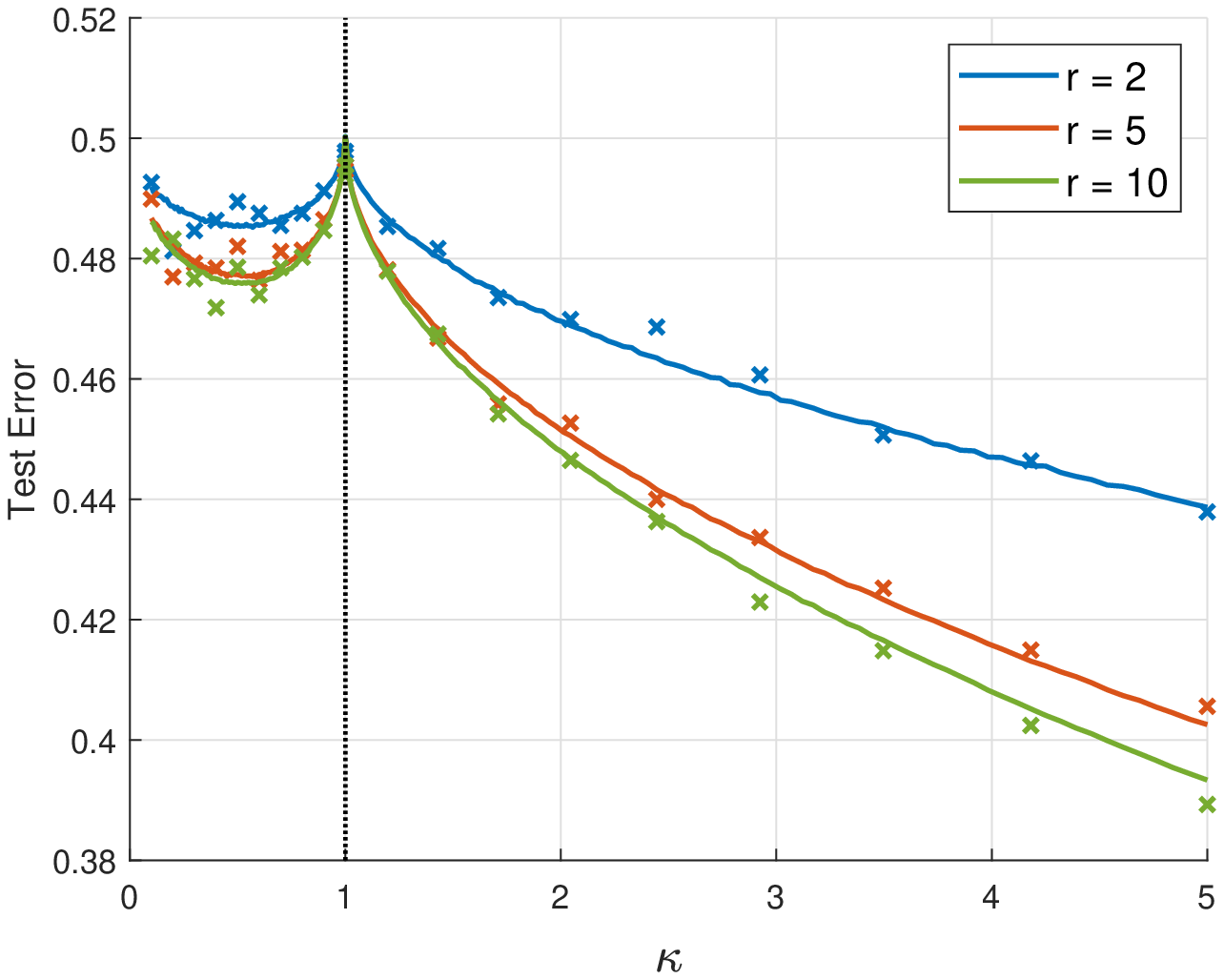}
		\caption{Linear feature selection rule.}
		\label{fig:log_lin}
	\end{subfigure}\hspace{0.5cm}%
	\begin{subfigure}{.48\textwidth}
		\centering
		\includegraphics[width=\linewidth]{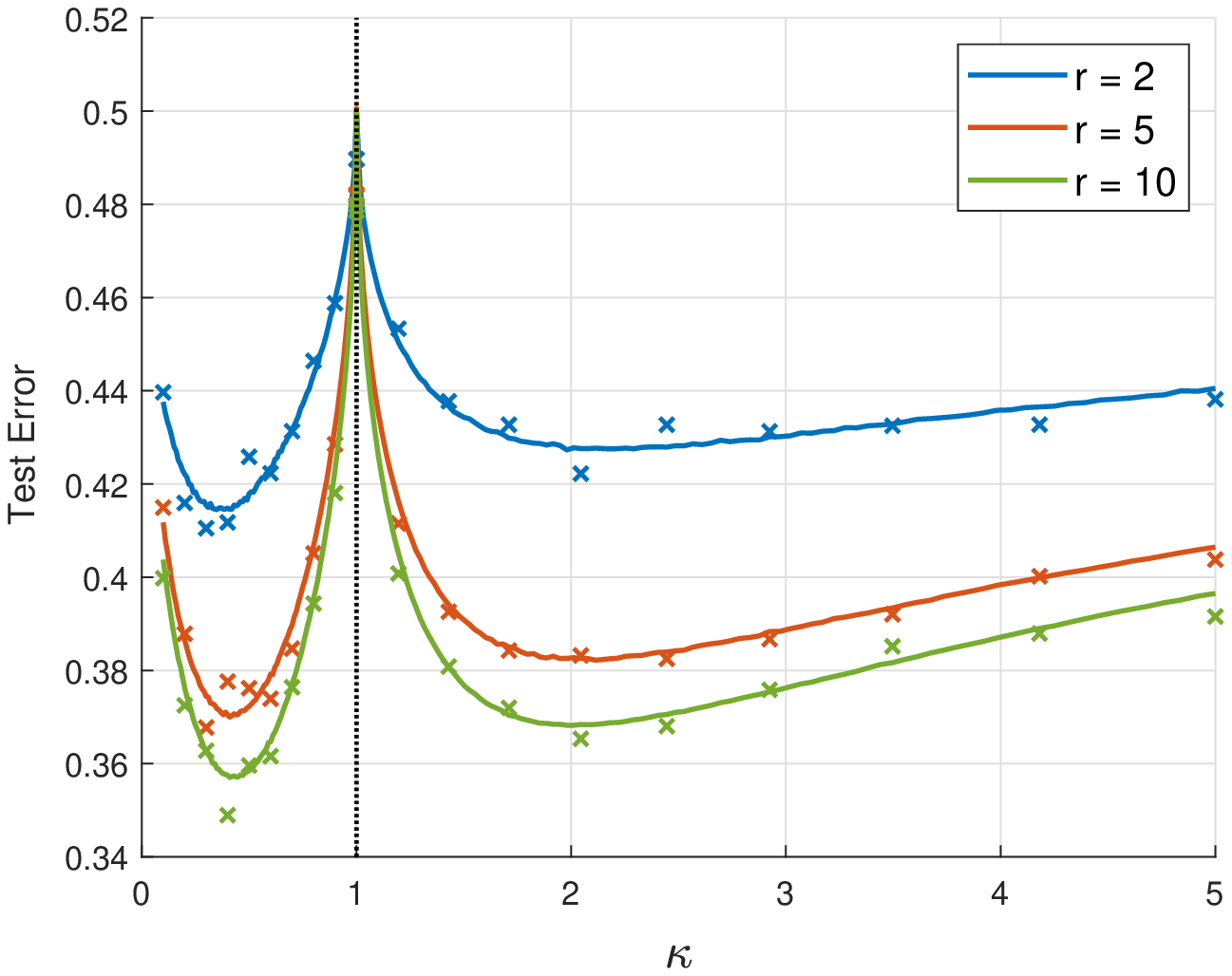}
		\caption{Polynomial feature selection rule with $\gamma=2$.}
		\label{fig:log_poly}
	\end{subfigure}
	\caption{Test error as a function of the overparametrization ratio $\kappa$ under a Logistic data model with three different SNR values $r=1,2$ and $5$. Simulation results are shown with crosses (`$\times$') and theoretical predictions with solid lines.} 
	\label{fig:log}
\end{figure}

\section{Numerical results \& Discussion}\label{sec:num}

\subsection{Feature selection models}\label{sec:explicit}

Results and discussions that follow are for two feature selection models. In linear regression setting, similar models are considered in \cite{breiman1983many,hastie2019surprises,belkin2019two,Zeyu}.


%
%
%

\vspace{5pt}
\noindent{\textbf{Linear model}:}~The signal strength $s^2=\|\betab_0\|_2^2$ increases linearly with the number $p$ of features considered in training. Specifically, for fixed $r^2$ and $\zeta>0$,
\bea\label{eq:uniform}
s^2 = s^2(\kappa) = r^2\cdot({\kappa}/{\zeta}),\quad\kappa\in(0,\zeta].
\eea
 This models a setting where all coefficients of the regressor $\etab_0$ have equal contribution. 

\vspace{5pt}
\noindent{\textbf{Polynomial model}:}~In contrast with the linear model, where the signal strength increases linearly by adding more features during training, the second model captures diminishing returns:
%
%
\bea\label{eq:poly}
s^2 = s^2(\kappa) = r^2\cdot\big(1-(1+\kappa)^{-\gamma}\big),~\kappa>0,
\eea
for some $\gamma\geq 1$.The signal strength $s$ increases with $\kappa$, but the increments  reduce in size with $\kappa$, as decided by $\gamma$.

\subsection{Risk curves}\label{sec:curves}

\begin{figure}
	\centering
	\begin{subfigure}{.48\textwidth}
		\centering
		\includegraphics[width=\linewidth]{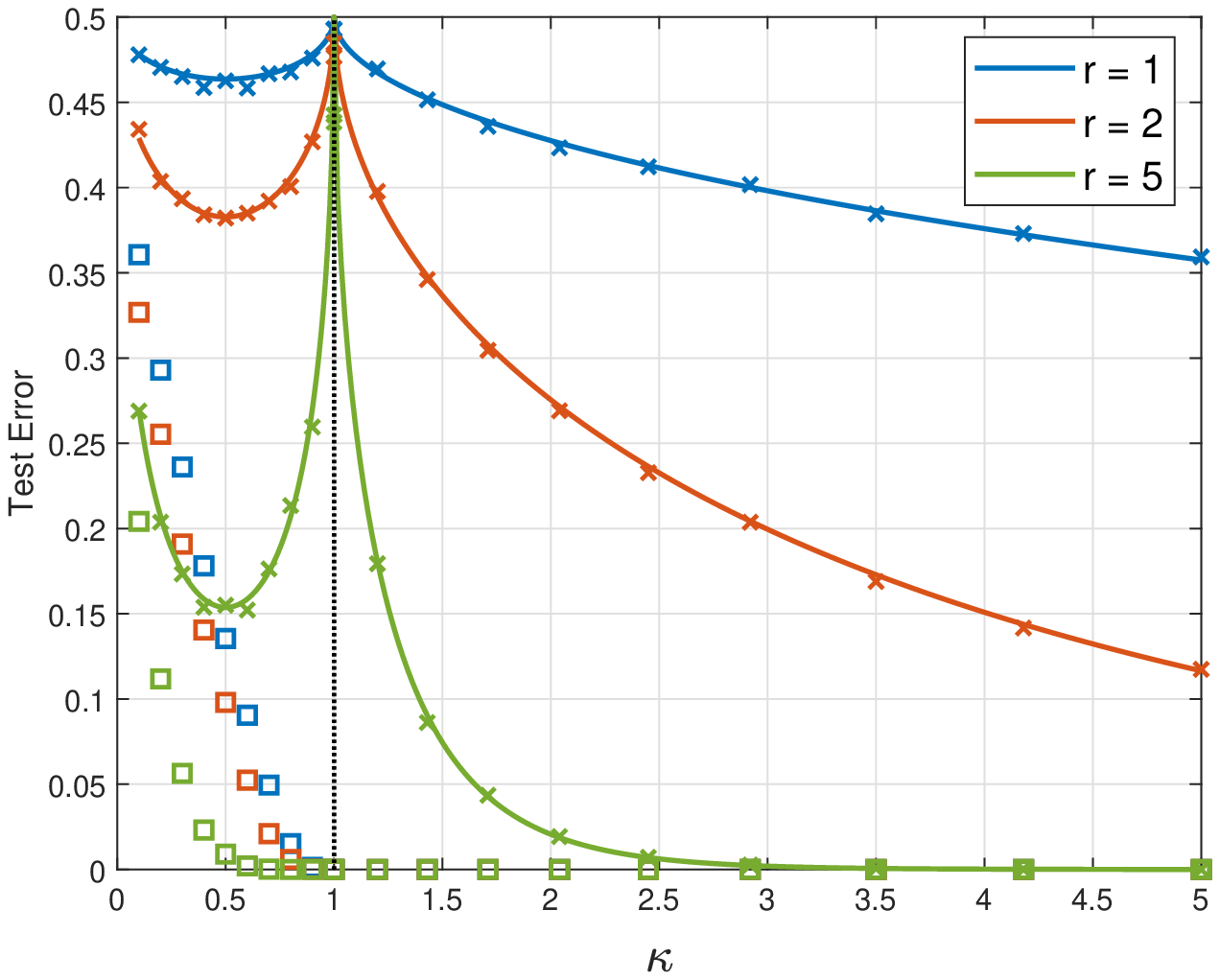}
		\caption{Linear feature selection rule.}
				\label{fig:gmm_lin}
	\end{subfigure}\hspace{0.5cm}%
	\begin{subfigure}{.48\textwidth}
		\centering
		\includegraphics[width=\linewidth]{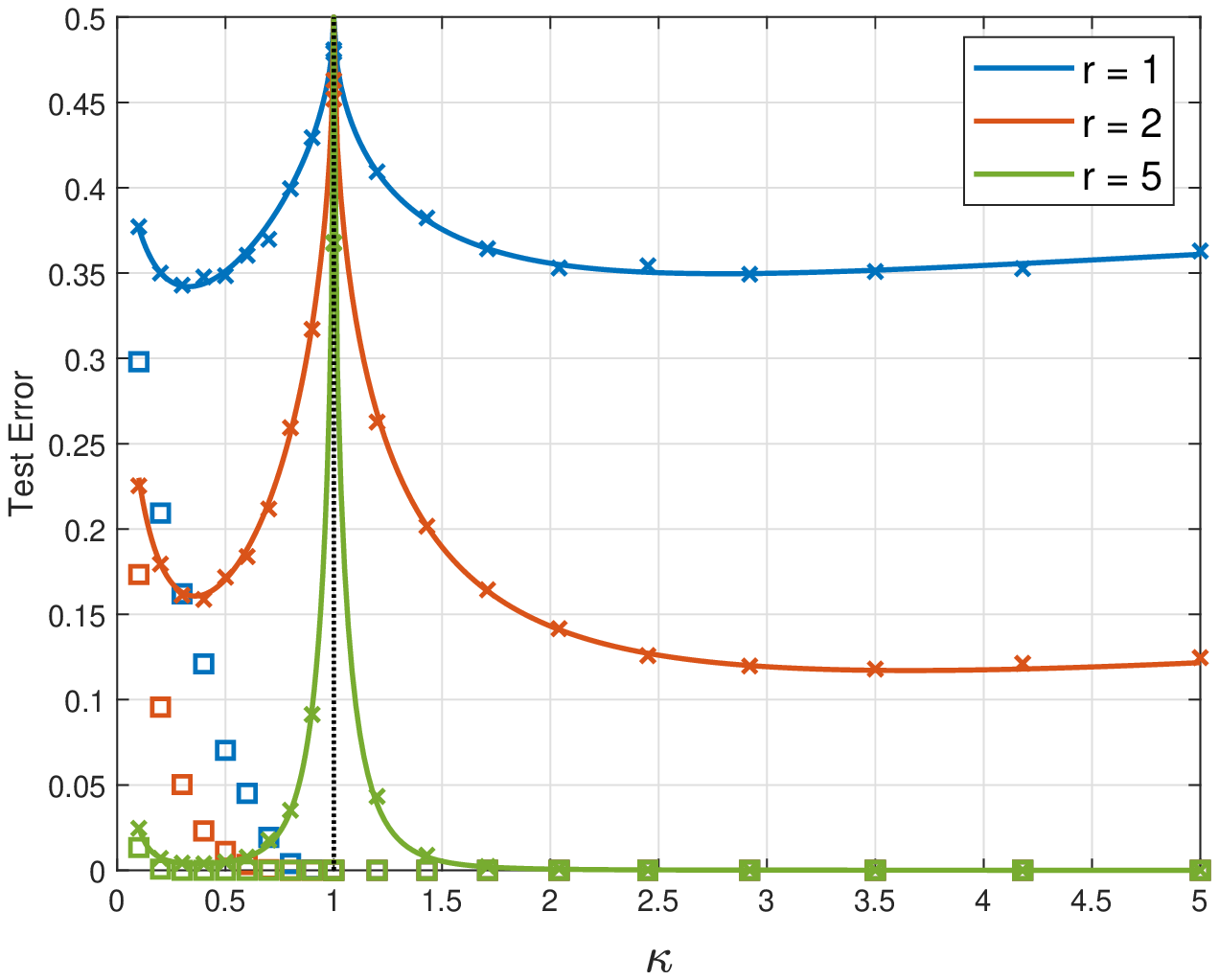}
		\caption{Polynomial feature selection rule with $\gamma = 2$.}
				\label{fig:gmm_poly}
	\end{subfigure}
	\caption{Test error as a function of the overparameterization ratio $\kappa$ under a GM data model with three different SNR values $r=1,2$ and $5$. Simulation results are shown with crosses (`$\times$') and theoretical predictions with solid lines. The square (`$\square$') datapoints depict empirical values of the training error.} 
	\label{fig:gmm}
\end{figure}

Figures \ref{fig:gmm_poly} and \ref{fig:log_poly} assume the polynomial feature model \eqref{eq:poly} for $\gamma=2$ and three values of $r$. Figures \ref{fig:gmm_lin} and \ref{fig:log_lin} show results with the linear feature model \eqref{eq:uniform} for $\zeta = 5$ and three values of $r$. Simulation results are shown with crosses (`$\times$'), and theoretical predictions with solid curves. Square-markers show training error, however these have been omitted in few plots to display test-error behavior more clearly. The dotted vertical lines depict the location of the linear interpolation threshold ($\kappa = 1$) according to \ref{propo:PT}. Simulation results shown here are for $n=500$ and the errors are averaged over $300$ independent problem realizations. Recall from \eqref{eq:train_error} that the training error is defined as the proportion of mismatch between the training set labels and the labels predicted by the trained regressor on the training data itself. Note that the phase-transition threshold of the least-squares does not coincide with value of $\kappa$ where the training error vanishes. However, as argued in Section \ref{sec:PT}, the training error is always zero for $\kappa>1$. In fact, the threshold $\kappa=1$ matches with the threshold where the empirical square loss (over the training set) becomes zero.

Figures \ref{fig:log_data_sq_loss_log_loss} and \ref{fig:gmm_data_sq_loss_log_loss} show a comparison between the predicted performance of the two algorithms, namely GD on square-loss and GD on logistic-loss, with logistic and GM data models, respectively. 
Finally, Figure \ref{fig:varying_n} demonstrates the predicted test error  as a function of $\frac{\kappa}{\zeta}$ for three values of $\zeta$.   

Overall, the simulations validate the theoretical predictions of Propositions \ref{propo:log} and \ref{propo:GM}. Experimental training error behavior also matches with the result of Proposition \ref{propo:PT}. The training error is zero with high probability when the training data can be linearly interpolated using the trained regressor under either of the two loss functions. This happens in the regime when $\kappa$, our measure of the model complexity, exceeds the phase-transition threshold of $\kappa = 1$.

\begin{remark}[Test error at high-SNR: logistic vs GM model] Comparing the curves for $r=5$ in Figures \ref{fig:log} and \ref{fig:gmm}, note that the test error under the GM model takes values (much) smaller than those under the logistic model. While the value of $r$ plays the role of SNR in both cases, the two models \eqref{eq:yi_log} and \ref{eq:yi_GM} are rather different. The discrepancy between the observed behavior of the test error can be understood as follows. On the one hand, in the GM model (cf. \eqref{eq:yi_GM}) the features satisfy  $\x_i=y_i\etab_0+\eps_i$, $\eps_i\sim\Nn(0,1)$. Thus, learning the vector $\etab_0$ involves a \emph{linear} regression problem with SNR $r=\|\etab_0\|_2$. On the other hand, in the logistic model (cf. \eqref{eq:yi_log}) at high-SNR $r\gg 1$ it holds $y_i\approx \sign(\x_i^T\etab_0)$. Hence, learning $\etab_0$ involves solving a system of \emph{one-bit} measurements, which is naturally more challenging than linear regression.
\end{remark}

\subsection{Discussion on double-descent}
The double-descent behavior of the test error as a function of the model complexity can be clearly observed in all the plots described above. Note that, the test error curve always has a U-shape in the underparametrized regime. As such, a first local minimum is always in the interior $(0,1)$. In contrast, in the overparametrized regime, the shape of the curve depends on the data. As an example, compare the risk curves for $\kappa>1$ for two different feature selection models in Figures \ref{fig:log_poly} and \ref{fig:log_lin}. The risk has a U-shaped curve in the first case, but is monotonically decreasing in the second. By further comparing (say) Figure \ref{fig:gmm_poly} to Figure \ref{fig:log_poly} the exact shape depends also on the data generation model and on the SNR. Perhaps more important is the related observation that the location of the global minimum of the risk curve (i.e., best performance for any $\kappa>0$) is also dependent on the data generation model and on the SNR. For example, the global minimum under the Logistic model in Figure \ref{fig:log_poly} is attained for $\kappa<1$ for all values of the SNR. In contrast, the best performance is attained in the overparametrized regime for data from the GM model and high-enough SNR.

\vp
\noindent\textbf{On the phase-transition threshold:}~Recall from Proposition \ref{propo:PT} that, for LS, the threshold distinguishes between two regimes: one where data can be linearly interpolated with a regressor of  size $p=\kappa\,n$, and one where it cannot. For the logistic loss, the distinction amounts to whether the data is linearly separable or not. In both cases, for model size $p$ larger than what the threshold determines, the empirical loss $\Rce$ over the training set is zero. This threshold distinguishes between the two U-shapes in the test-error curve. Our results show that this threshold depends in general on both the data (through the models of data -- logistic or GM -- and through the SNR) and on the training algorithm, which in our setting is determined by the choice of the loss function in \eqref{eq:Remp}.

\begin{figure}
	\centering
	\includegraphics[width=.48\linewidth]{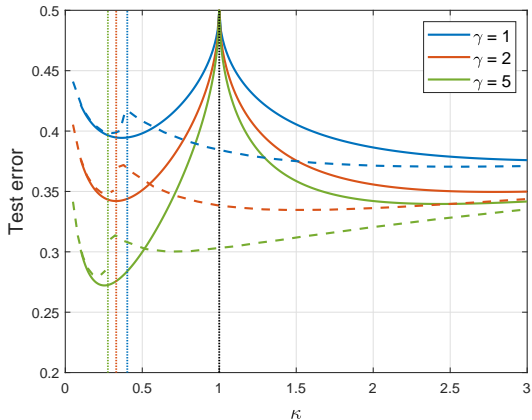}
	\caption{The effect of the training algorithm (square- vs logistic- loss) on the DD curves of the test error as a function of the overparametrization ratio $\kappa$ for the polynomial feature selection model under GM data. Risk curves for square-loss are shown as solid lines and those for logistic-loss are shown as dashed lines. The vertical lines indicate the phase-transition threshold for each case.}
	\label{fig:gmm_data_sq_loss_log_loss}
\end{figure}

\vp
\noindent\textbf{Logistic vs square loss:} In Figures \ref{fig:log_data_sq_loss_log_loss} and \ref{fig:gmm_data_sq_loss_log_loss}, we plot the test error of GD with both square and logistic loss for three instances of the polynomial feature selection rule under the logistic model and the GM model, respectively. For the square loss, we use the results of Proposition \ref{propo:log}, while the curves for logistic-loss follow from \cite[Prop.~3.2]{Zeyu} and \cite[Prop.~3.3]{Zeyu}. In both cases, the test error exhibits a clear double-descent behavior. However, our comparative study reveals important differences in the following features of these curves:

\vspace{1pt}
\noindent$\bullet$~\emph{Cusp of the curve:}~It becomes apparent that the location of the cusp of the curves depends on the algorithm, i.e., the loss function being minimized. On the one hand, for least-squares, the cusp is always at $\kappa=1$, i.e., at the threshold after which the data can be linearly interpolated. On the other hand, the threshold for logistic loss is at $\kappa_{\star,\rm LOG}$, which is data-dependent and determines the value after which the data are linearly separable \cite[Prop.~3.1]{Zeyu}. 

\vspace{1pt}
\noindent$\bullet$~\emph{Global minimum:} For $\kappa<\kappa_{\star,\rm LOG}$ the test error of LS and logistic are almost indistinguishable. This fact is theoretically justified in \cite{AISTATS2020}, where it is shown that the square loss is (approximately) optimal for binary logistic classification in the regime $p<n$, among all choices of losses for which the optimal set in \eqref{eq:Remp} is finite. Instead, when $p>n$, we observe in Figures \ref{fig:log_data_sq_loss_log_loss} and \ref{fig:gmm_data_sq_loss_log_loss} that the logistic loss is always better. However, note that the global minimum of the curves in the entire range $\kappa>0$, depends on the data SNR. For instance, in Figure \ref{fig:log_data_sq_loss_log_loss}, for $\gamma=2$ the minimum of the curve corresponding to Logistic loss is below that of the LS curve, but for $\gamma=5$ the opposite is true. 

\vspace{1pt}
\noindent$\bullet$~\emph{Stability to model selection perturbations:} Finally, it is worth observing that the test error for logistic loss is less sensitive to variations of the model size $p$, even around the threshold $\kappa_{\star,\rm LOG}$. This is to be contrasted to the LS curve, which undergoes sharp changes (sudden increase and decrease) close to the linear interpolation threshold $\kappa=1$. 
\begin{figure}
		\centering
		\includegraphics[width=0.5\linewidth]{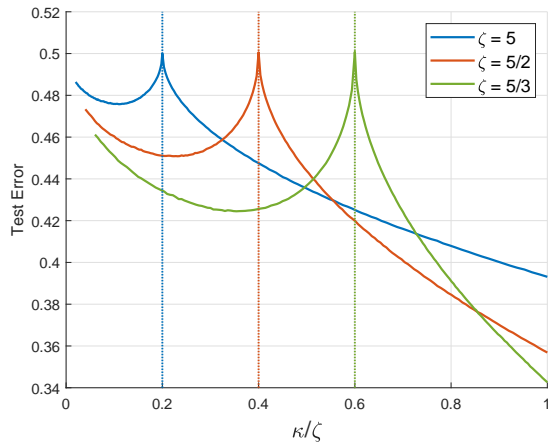}
		\caption{The effect of increasing the training size $n$ on the DD curves. Different values of $\zeta$ correspond to different training sizes: $n=\zeta d$. The x-axis has been normalized so that it does not depend on $n$; indeed, recall that $\kappa/\zeta=p/d$.  Test error is plotted for the linear feature selection model (cf. \eqref{eq:uniform}) under the Logistic data model. The vertical lines depict the linear interpolation threshold ($p=n$). Larger training-sample sizes increase the interpolation threshold as predicted by the theory. Also, due to the sharp changes of the error close to the threshold, there are choices of model size $p$ for which more data can hurt; see text for details.}
		\label{fig:varying_n}
\end{figure}

\vp
\noindent\textbf{On the size of the training set:} In Figure \ref{fig:varying_n}, we investigate the dependence of the double-descent curves on the size of the training set. Specifically, we fix $d$ (the dimension of the ambient space) and study DD for three distinct values of the training-set size, namely $n=\frac{1}{5}d, \frac{2}{5}d,$ and $\frac{3}{5}d$. For these three cases and the linear feature-selection model, we plot the test error vs the model size $p=\frac{\kappa}{\zeta} d$. First, observe that as the training size grows larger (i.e., $\zeta$ increases), the interpolation threshold shifts to the right (thus, it also increases). Second, while larger $n$ naturally implies that the global minima (here, at $p=d$ for all $n$) of the corresponding curve corresponds to better test error, the curves cross each other. This means that more training examples do not necessarily help, and could potentially hurt the classification performance, whether at the underparametrized or the overparametrized regimes. For example, for model size $p\approx 0.5 d$, the test error is lowest for $n=\frac{1}{5}d$ and largest for $n=\frac{3}{5}d$. The effect of the size of the training set on the DD curve was recently studied \emph{empirically} in \cite{DDD}; see also \cite{nakkiran2019more} for an analytical treatment in a linear regression setting. Our investigation is motivated by and theoretically corroborates the observations reported therein. 

\section{Future work}
In this and the companion paper\cite{Zeyu}, we study double-descent curves in binary linear classification under simple models with Gaussian features. The proposed setting is simple enough that it allows a principled analytic study of several important features of DD curves. Specifically, we investigate the dependence of the curve's transition threshold and global minimum on: (i) the loss function (aka training procedure); (ii) the learning model and SNR; and, (iii) the size of the training set. Throughout, we emphasize the resemblance of our conclusions to corresponding empirical findings in the literature \cite{DDD,belkin2018reconciling,spigler2019jamming}.

We believe that this line of work can be extended in several aspects and we briefly discuss a few of them here. To begin with, it is important to better understand the effect of correlation on the DD curve. Note that simple feature selection rules such as those of Section \ref{sec:explicit} can only model covariance matrices $\Sigma=\E[\x\x^T]$ that are diagonal. The papers \cite{hastie2019surprises} and \cite{montanari2019generalization} derive sharp asymptotics on the performance of min-norm estimators under correlated Gaussian features for linear regression and linear classification, respectively. While this allows to numerically plot DD curves, the structure of $\Sigma$ enters the asymptotic formulae through unwieldy expressions. Thus, understanding how different correlation patterns affect DD to a level that may provide practitioners with useful insights and easy take-home-messages remains a challenge; see \cite{bartlett2019benign} for related efforts. Another important extension is that to multi-class settings. The recent work \cite{DDD} includes a detailed empirical investigation of the dependence of DD on label-noise and data-augmentation. Moreover, the authors identify that ``DD occurs not just as a function of model size, but also as a function of training epochs". It would be enlightening to analytically study whether these type of behaviors are already present in simple models similar to those considered here.  Yet another important task is carrying out the analytic study to more complicated --nonlinear-- data models; see \cite{mei2019generalization,montanari2019generalization} for progress in this direction.


%
%
%

\bibliography{compbib}


%
%
%
%
%

\end{document}